\documentclass[conference]{IEEEtran}
\IEEEoverridecommandlockouts
\usepackage{cite}
\usepackage{amsmath,amssymb,amsfonts}
\usepackage{algorithmic}
\usepackage{graphicx}
\usepackage{textcomp}
\usepackage{siunitx}
\usepackage{tabularx}
\usepackage{xcolor}
\usepackage{subcaption}
\usepackage[font=small]{caption}
\usepackage{wrapfig}
\usepackage{booktabs}
\usepackage{multirow}
\usepackage{algorithm}
\usepackage{tcolorbox}  
\usepackage{url}
\newcommand{\bluetext}[1]{\textcolor{blue}{#1}}

\newcommand{\MS}[1]{\textcolor{red}{#1}}

\def\BibTeX{{\rm B\kern-.05em{\sc i\kern-.025em b}\kern-.08em
    T\kern-.1667em\lower.7ex\hbox{E}\kern-.125emX}}

\begin{document}

\title{TinyML NLP Scheme for Semantic Wireless Sentiment Classification with Privacy Preservation}

\author{
	\IEEEauthorblockN{Ahmed Y. Radwan$^1$, Mohammad Shehab$^2$, and Mohamed-Slim Alouini$^2$\\
	}
	\IEEEauthorblockA{$^1$Department of Electrical Engineering and Computer Science, York University, Toronto, ON M3J 1P3, Canada\\
$^2$CEMSE Division, King Abdullah University of Science and Technology (KAUST), Thuwal 23955-6900, Saudi Arabia\\  
Email: ahmedyra@yorku.ca
    }
    }

\maketitle

\begin{abstract}
Natural Language Processing (NLP) operations, such as semantic sentiment analysis and text synthesis, often raise privacy concerns and demand significant on-device computational resources. Centralized learning (CL) on the edge provides an energy-efficient alternative but requires collecting raw data, compromising user privacy. While federated learning (FL) enhances privacy, it imposes high computational energy demands on resource-constrained devices. This study provides insights into deploying privacy-preserving, energy-efficient NLP models on edge devices. We introduce semantic split learning (SL) as an energy-efficient, privacy-preserving tiny machine learning (TinyML) framework and compare it to FL and CL in the presence of Rayleigh fading and additive noise. Our results show that SL significantly reduces computational power and CO\textsubscript{2} emissions while enhancing privacy, as evidenced by a fourfold increase in reconstruction error compared to FL and nearly eighteen times that of CL. In contrast, FL offers a balanced trade-off between privacy and efficiency. Our code is available for replication at our GitHub repository: \url{https://github.com/AhmedRadwan02/TinyEco2AI-NLP}.

\end{abstract}


\section{Introduction}

Artificial Intelligence (AI) has gained widespread adoption, with applications ranging from text and image classification to text and image generation. NLP has seen rapid growth, driving advancements in virtual assistants and language classification systems. Transformer-based Large Language Models (LLMs), such as OpenAI’s GPT series \cite{gpt4_technical_report} and BERT, have significantly advanced NLP by enabling machines to perform complex tasks with high accuracy. However, these models require substantial computational resources for training and inference, making them less practical for deployment on resource-constrained devices. Additionally, older temporal models like long short-term memory networks (LSTMs) and other recurrent neural network (RNN) variants remain a popular choice due to their ability to balance computational efficiency and performance.

Although more efficient than LLMs, temporal models still require substantial storage and processing power. The high dimensionality of language data and large vocabularies further amplify the computational demands \cite{lin2023tiny}.

Privacy is another critical concern, as training robust models often necessitates diverse datasets, potentially exposing sensitive information. While CL exacerbates this risk by collecting raw data, FL mitigates it through decentralized training. However, FL remains vulnerable to inference attacks on gradient updates. SL addresses this vulnerability by transmitting only intermediate activations, significantly reducing the risk of data reconstruction.

In addition to privacy and computation, edge deployment introduces further challenges. Limited processing capacity, bandwidth constraints, and latency issues can hinder model performance. Moreover, wireless data transmission over channels such as WiFi is particularly susceptible to noise, fading, and unstable connections, which can degrade communication efficiency.

TinyML mixed models have evolved to address the above concerns. For instance, model compression techniques, such as pruning \cite{reed1993pruning}, quantization \cite{quantization}, and knowledge distillation \cite{knowledge_distillation_survey}, help reduce model sizes and computational requirements. TinyBERT \cite{tinybert}, for example, utilizes knowledge distillation to create smaller models, while quantization techniques, like in LLaMA \cite{touvron2023llama}, enable efficient execution on low-resource devices. FL offers a decentralized model training approach, allowing multiple users to train a global model while keeping their data local collaboratively. 
To this end, SL \cite{vepakomma2018split} divides the model training process between users and a server. Users transmit activations from the initial model layers, reducing computational load {while further limiting raw data exposure}. 

While TinyFedTL \cite{kopparapu2022tinyfedtl} introduced FL for resource-constrained microcontrollers and TDMiL \cite{gulati2024tdmil} designed a framework for distributed learning in microcontroller networks, both approaches fall short of addressing key challenges in decentralized NLP. TinyFedTL focuses on federated transfer learning, omitting the exploration of alternative learning paradigms like SL and lacks considerations for wireless channel effects. Similarly, TDMiL primarily addresses computational variability and synchronization but relies on wired setups and overlooks the unique challenges of over-the-air communication. In contrast, our work bridges these gaps by proposing and evaluating decentralized learning frameworks for text emotion classification. Moreover, the extraction of information semantics and transmitting it instead of the information itself would significantly save bandwidth and reduce energy consumption \cite{yang2022semantic}. 

Unlike TinyFedTL and TDMiL, we introduce realistic wireless channel impairments, such as Rayleigh fading and additive noise, into the learning process. This inclusion allows for a more comprehensive evaluation of decentralized learning in real-world scenarios. Furthermore, our semantic SL-based approach significantly reduces computational power and CO\textsubscript{2} emissions by using a partitioned model architecture, with only the initial layers processed on the device. {We further assess privacy preservation across FL, SL, and CL by measuring reconstruction loss, demonstrating SL’s superior resistance to data leakage.} These contributions not only extend the applicability of decentralized learning to wireless NLP applications but also align with TinyML initiatives by enhancing resource efficiency and sustainability.

The remainder of this paper is structured as follows: Section~II covers the system design and experimental setup, including both frameworks of FL and SL techniques. Section~III presents the experimental results, focusing on accuracy, energy consumption, {privacy evaluation}, and the impact of noise and fading. Finally, Section~IV provides conclusions and discusses potential directions for future work.

\begin{figure*} [h!]
    \centering
    \includegraphics[width=0.92\linewidth, trim=20 15 20 10, clip]{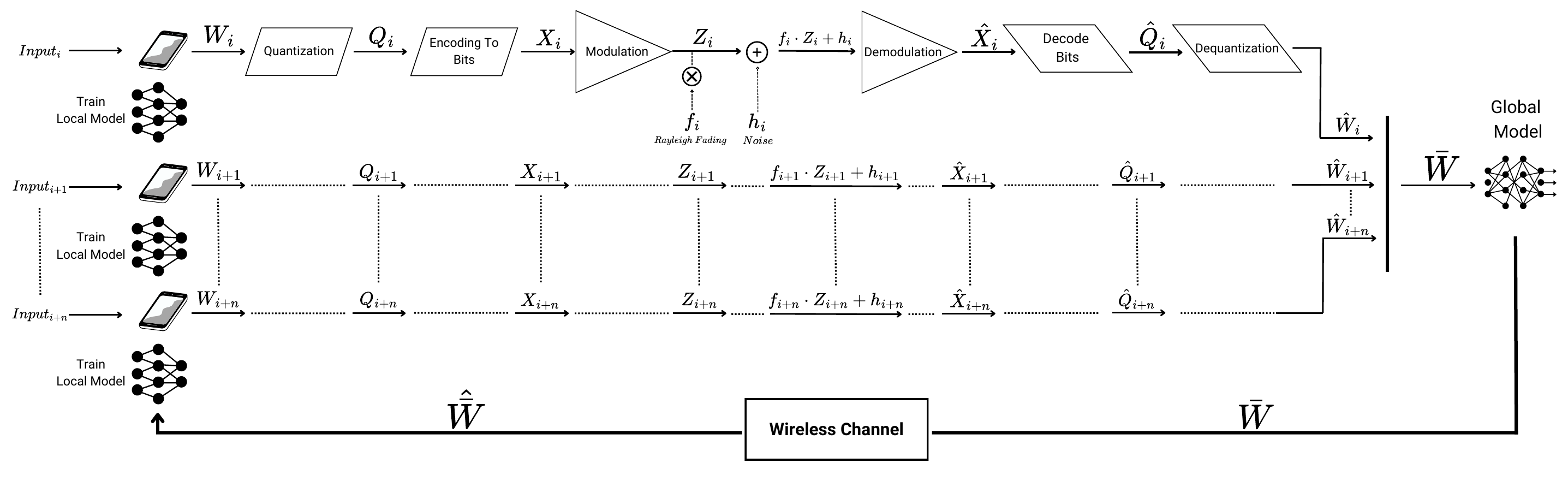}
    \caption{\small{Federated learning cycle: users train on local data, send model updates to server for aggregation}}
    \label{fig:federatedSystemDesign} 
\end{figure*}

\section{System Layout}\label{sec:sys}
    


\subsection{Federated Learning System}

As shown in Fig \ref{fig:federatedSystemDesign}, in FL, $N$ users collaborate by training local models and sharing their updates with a central server over $K$ communication cycles. The key operation in FL is the transmission of quantized model updates to minimize communication overhead and ensure data privacy, as raw data never leaves the users' devices.
In each communication cycle $k$, user $i$ performs $J$ local training iterations on their private dataset to update model weights $W_i^{(k)}$. This iterative local training allows users to improve their models before sharing updates with the server.
The updated local model weights $W_i^{(k)}$ are quantized to reduce communication overhead. The quantization process converts the weights into discrete levels, scaled by a factor $S_i^{(k)}$ derived from the maximum absolute weight value and the bit-width $b$ used for quantization. The quantized weights are computed as 
\begin{equation}\label{eq:quantize_weights}
Q_i^{(k)}=\left\lceil \frac{W_i^{(k)}}{S_i^{(k)}} \right\rceil,
\end{equation}
\vspace{-1mm}  where the scale factor $S_i^{(k)}$ is defined as $S_i^{(k)} = \frac{\max\left(|W_i^{(k)}|\right)}{2^{b-1} - 1}$.

The quantized weights $Q_i^{(k)}$ are then encoded into a bit stream $X_i^{(k)}$, which is then digitally modulated into a signal $Z_i^{(k)}$. This signal is transmitted through the wireless channel. Consequently, the received signal $\hat{Z}_i^{(k)}$ is a faded and noisy version of the originally transmitted signal due to Rayleigh fading $f_i^{(k)}$ and additive noise $n_i^{(k)}$. These channel effects can lead to slight changes in the demodulated bit stream $\hat{X}_i^{(k)}$ \cite{xiao2024overfadingchannel}.
The server then collects the received updates $\hat{Z}_i^{(k)}$ from each user $i$. These signals are first demodulated into the bit stream $\hat{X}_i^{(k)}$, which is then decoded into the quantized weights $\hat{Q}_i^{(k)}$. After decoding, the server dequantizes $\hat{Q}_i^{(k)}$ to recover the estimated model weights $\hat{W}_i^{(k)}$
\begin{equation}
    \label{eq:dequantize_weights}
    \hat{W}_i^{(k)} = \hat{Q}_i^{(k)} \cdot S_i^{(k)}.
\end{equation}
To obtain the global model update, the server aggregates these dequantized updates using Federated Averaging (FedAvg)~\cite{nilsson2018performancefedavg} as
\begin{equation}
    \label{eq:fed_avg}
    \bar{W}^{(k+1)} = \frac{1}{N} \sum_{i=1}^{N} \hat{W}_i^{(k)}.
\end{equation}

Here, $\bar{W}^{(k+1)}$ represents the updated global model after aggregating the users' contributions. The server then broadcasts $\bar{W}^{(k+1)}$ back to the users, who update their local models accordingly for the next cycle \vspace{-1mm}

\begin{equation}
    \label{eq:local_model_update}
    W_i^{(k+1)} = \bar{W}^{(k+1)}.
\end{equation}
This process is repeated for $K$ communication cycles, allowing the global model to converge through iterative refinement while maintaining data privacy. The whole FL process is summarized in Algorithm \ref{alg:efficient_federated_learning}.

\subsection{Split Learning System}

\begin{figure*}[h!]
    \centering
    \includegraphics[width=0.88\linewidth, trim=20 20 20 20, clip]{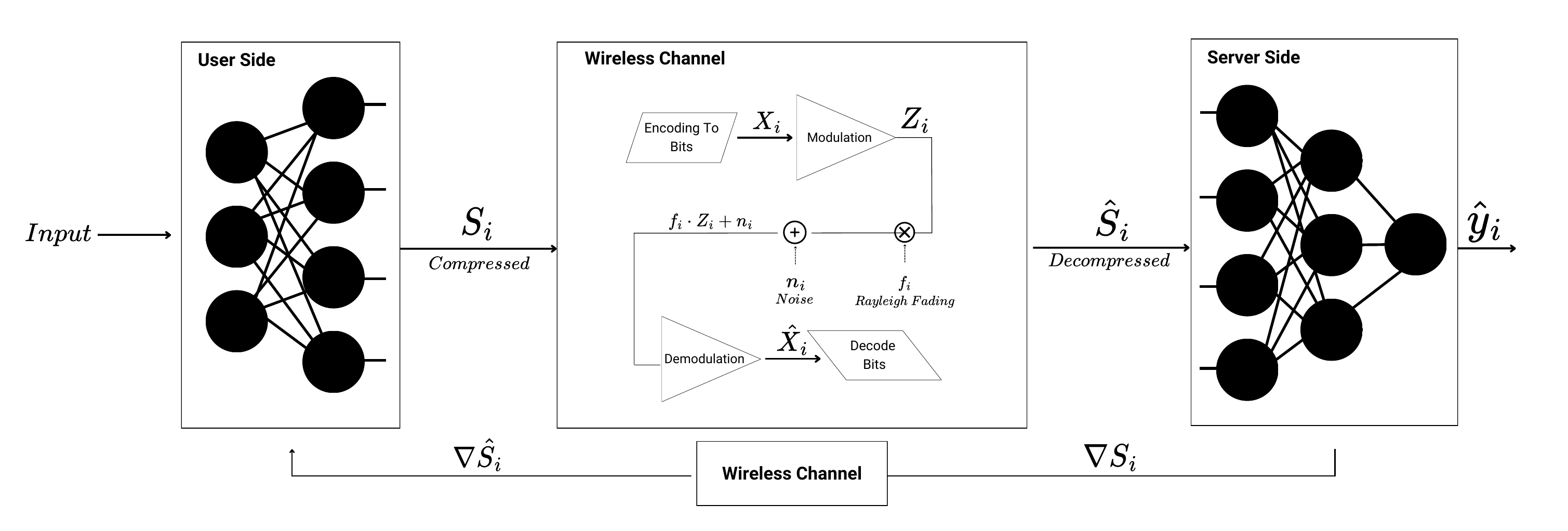} 
    \caption{\small {Split learning system design for one user. The user processes layers and sends activations to the server for processing.}}
    \label{fig:splitSystemDesign}
\end{figure*}

SL enables a distributed learning approach where the model is divided between a user and a server, as shown in Fig.~\ref{fig:splitSystemDesign}. In each communication cycle $k$, the user processes only the initial layers of the model locally, which reduces the computational burden and protects raw data privacy. Then, it transmits the intermediate activations to the server for further processing. Specifically, for a given input $x_i$ at cycle $k$, the user computes the smashed data
\begin{equation}
    \label{eq:user_output}
    S_i^{(k)} = f_{\text{user}}(x_i; W_{\text{user}}^{(k)}),
\end{equation}
where $W_{\text{user}}^{(k)}$ are the user-side model weights at cycle $k$, and $S_i^{(k)}$ represents the output of the user's model partition (i.e., smashed data).
The output $S_i^{(k)}$ is first encoded into bits $X_i^{(k)}$ and then digitally modulated into a signal $Z_i^{(k)}$. This signal is transmitted through the wireless channel as will be defined in Section~\ref{sec:wireless_channel_model}.
Due to channel effects, errors can be introduced in the received signal, affecting the accuracy of the demodulated bit stream $\hat{X}_i^{(k)}$.
Upon receiving the noisy signal, the server demodulates and decodes it to obtain the estimated activations $\hat{S}_i^{(k)}$, which are then utilized to complete the forward pass

\begin{equation}
    \label{eq:server_output}
    \hat{y}_i^{(k)} = f_{\text{server}}(\hat{S}_i^{(k)}; W_{\text{server}}^{(k)}),
\end{equation}
where $W_{\text{server}}^{(k)}$ are the server-side model weights at cycle $k$, and $\hat{y}_i^{(k)}$ is the server's output prediction. The loss function is computed as
\begin{equation}
    \label{eq:loss_function}
    L^{(k)} = \mathcal{L}(\hat{y}_i^{(k)}, y_i),
\end{equation}
where $y_i$ is the ground truth label, and $\mathcal{L}$ denotes the loss function (e.g., cross-entropy loss).


The server computes gradients with respect to its model weights and the activations. The gradient $\nabla_{\hat{S}_i^{(k)}} L^{(k)}$ is clipped to manage its magnitude and then transmitted back to the user through the wireless channel. The clipped gradient is encoded into bits, modulated, and subjected to channel impairments, similar to the forward transmission.
The server updates its model weights using

\begin{equation}
    \label{eq:server_weight_update}
    W_{\text{server}}^{(k+1)} = W_{\text{server}}^{(k)} - \eta \nabla_{W_{\text{server}}^{(k)}} L^{(k)},
\end{equation}
where $\eta$ is the learning rate.
The user receives the noisy version of the gradients, demodulates and decodes them to obtain the estimated gradients $\widehat{\nabla}_{\hat{S}_i^{(k)}} L^{(k)}$, and then performs backpropagation on the local layers to compute $\nabla_{W_{\text{user}}^{(k)}} L^{(k)}$. The user updates its model weights as

\begin{equation}
    \label{eq:user_weight_update}
    W_{\text{user}}^{(k+1)} = W_{\text{user}}^{(k)} - \eta \nabla_{W_{\text{user}}^{(k)}} L^{(k)}.
\end{equation}
This process is repeated over $K$ communication cycles, allowing both the user and server models to improve, iteratively. The process of SL is summarized in Algorithm \ref{alg:efficient_split_learning}.

\subsection{Wireless Channel} \label{sec:wireless_channel_model}
In both learning systems, semantically encoded model updates or activations (denoted as \( Z \)) are transmitted over wireless channels subject to Rayleigh fading and additive white Gaussian noise (AWGN) $n$. The fading coefficient \( f \) uniformly affects all transmitted signals \( Z_i \), and the noisy transmission is represented by
\begin{equation}
    \label{eq:noisy_signal}
    \hat{Z} = f \cdot Z + n, 
\end{equation}
and similar definitions apply for the feedback in both cases.
\begin{algorithm}[t!]
\caption{Federated Learning for Semantic Wireless Text Sentiment Classification}
\label{alg:efficient_federated_learning}

\begin{algorithmic}[1]
\STATE \textbf{Initialize:} $\eta, J, K, \sigma^2, Q, N, \mathbf{\hat{W}}^{(0)}$
\FOR{$k = 1$ to $K$}
    \FOR{$i = 1$ to $N$}
        \STATE \textbf{User $i$:} $W_i^{(k)} = \mathbf{\hat{W}}^{(k)}$ 
        \FOR{$j = 1$ to $J$}
            \STATE $W_i^{(k)} \gets W_i^{(k)} - \eta \nabla L_i(W_i^{(k)})$
        \ENDFOR
        \STATE $Q_i^{(k)} = \text{Quantize}(W_i^{(k)}, Q)$ \COMMENT{Using \eqref{eq:quantize_weights}}
        \STATE $X_i^{(k)} = \text{Encode}(Q_i^{(k)})$
        \STATE $Z_i^{(k)} = \text{Modulate}(X_i^{(k)})$
        \STATE Transmit: $\tilde{Z}_i^{(k)} = f_i \cdot Z_i^{(k)} + n_i$, using  \eqref{eq:noisy_signal}
        
    \ENDFOR
    \STATE \textbf{Server:}
    \STATE $\hat{Z}_i^{(k)} = \text{Demodulate}(\tilde{Z}_i^{(k)})$ $\forall i$
    \STATE $\hat{X}_i^{(k)} = \text{Decode}(\hat{Z}_i^{(k)})$ $\forall i$
    \STATE $\hat{Q}_i^{(k)} = \text{Dequantize}(\hat{X}_i^{(k)})$ $\forall i$
    \STATE Obtain $\bar{W}^{(k+1)}$ using Using \eqref{eq:fed_avg} and broadcast it to back all users.
\ENDFOR
\end{algorithmic}
\end{algorithm}

\begin{algorithm}[h]

    \caption{Split Learning for Semantic Wireless Text Sentiment Classification}
    \label{alg:efficient_split_learning}
    \begin{algorithmic}[1]
        \STATE \textbf{Initialize:} $\eta, K, \sigma^2, L, N, \tau, W_{\text{user}}^{(0)}, W_{\text{server}}^{(0)}$
        
        \FOR{$k = 1$ to $K$}
            \FOR{$i = 1$ to $N$}
                \STATE \textbf{User $i$:}
                \STATE Compute $S_i^{(k)} \gets \text{UserOutput}(W_{\text{user}}^{(k)})$ \COMMENT{Eq. \eqref{eq:user_output}}
                \STATE Transmit $\tilde{Z}_i^{(k)}$ \COMMENT{Using \eqref{eq:noisy_signal}}
            \ENDFOR
            
            \STATE \textbf{Server:}
            \STATE Compute $\hat{y}_i^{(k)} \gets \text{ServerProcess}(\tilde{Z}_i^{(k)})$ \COMMENT{Decode, Demodulate; Using \eqref{eq:server_output}}
            \STATE Compute loss $L \gets \text{LossFunction}(\hat{y}_i^{(k)}, y_i)$ \COMMENT{Using \eqref{eq:loss_function}}
            \STATE Clip gradients $g_{\text{server}}^{\text{clipped}} \gets \text{clip}_{\text{norm}}(\nabla L_{\text{server}}, \tau)$
            \STATE Update $W_{\text{server}}^{(k+1)}$ \COMMENT{Eq. \eqref{eq:server_weight_update}}
            
            \STATE Compute gradients $\nabla S_i^{(k)} \gets \nabla \text{UserOutput}(W_{\text{user}}^{(k)})$
            \STATE Transmit gradient $\tilde{Z}_{\nabla i}^{(k)}$ \COMMENT{Using  \eqref{eq:noisy_signal}}

            \FOR{$i = 1$ to $N$}
                \STATE \textbf{User $i$:}
                \STATE Clip user gradients $g_{\text{user}}^{\text{clipped}} \gets \text{clip}_{\text{norm}}(\widehat{\nabla S_i^{(k)}} \cdot \nabla W_{\text{user}}^{(k)}, \tau)$
                \STATE Update $W_{\text{user}}^{(k+1)}$ \COMMENT{Using \eqref{eq:user_weight_update}}
            \ENDFOR
        \ENDFOR
    \end{algorithmic}
\end{algorithm}

\subsection{Communication Energy Calculation}
In our approach, communication energy is determined by calculating the channel capacity \( C \), which represents the maximum data rate for error-free transmission. Using the Shannon-Hartley theorem~\cite{shannon1949communication}, the channel capacity incorporates the Signal-to-Noise Ratio (SNR), multiplied by the Rayleigh fading $f$ and the bandwidth \( B \).
The time required to transmit one bit and the corresponding energy consumption per bit is derived from the following equations. The SNR represents the ratio of transmission power \( P \) to noise power \( \sigma^2 \).
After calculating the SNR we proceed to calculate the channel capacity \( C \) as 
\begin{equation}
C = B \log_2(1 + |f|^2 \text{SNR}).
\end{equation}
Finally, the energy consumed per transmitted bit is given by $\frac{P}{C}   (J/b),$
which is multiplied by the total number of bits (payload) to estimate the consumed communication energy.

\subsection{Privacy Evaluation}

{We evaluate privacy by assessing the ease of reconstructing raw input data from transmitted information.}
{The reconstruction error quantifies how much raw data can be recovered and is defined as:}
\begin{equation}
    Error = \frac{1}{N} \sum_{i=1}^N (x_i - \hat{x}_i)^2,
\end{equation} \vspace{-1mm}
where $x_i$ is the original input and $\hat{x}_i$ is the reconstructed data. In FL, $\hat{x}_i$ corresponds to transmitted weights $W_i$, while in SL, it refers to intermediate activations $S_i$.
{To improve privacy, normalization of the data is applied to avoid value spikes that might result in reconstruction easier.}

\section{Experiments and Results}

\begin{table}[h!]
    \centering
    \caption{Experimental Parameters}
    \small 
    \label{table:experiment_params}
    \newcolumntype{C}{>{\centering\arraybackslash}X}
    \begin{tabularx}{\linewidth}{l C C}
        \toprule
        \textbf{Parameter}               & \textbf{FL}                           & \textbf{SL}                           \\ 
        \midrule
        Training-Testing Split          & \multicolumn{2}{c}{90\% training, 10\% testing}           \\ 
        Vocabulary Size                 & \multicolumn{2}{c}{10,000 most frequent words}            \\ 
        Maximum Sequence Length         & \multicolumn{2}{c}{30}                                        \\ 
        Number of Users                 & 3                                     & 1                                      \\ 
        Communication Cycles            & 7 cycles, 5 epochs/user                & 50 cycles                              \\ 
        Batch Size                      & \multicolumn{2}{c}{512}                                       \\ 
        Optimizer                       & \multicolumn{2}{c}{SGD}                                       \\ 
        Momentum                        & \multicolumn{2}{c}{0.9}                                       \\ 
        Initial Learning Rate           & \multicolumn{2}{c}{0.01}                                      \\ 
        Learning Rate Scheduler         & \multicolumn{2}{c}{Reduce by 10\% every 5 epochs}             \\ 
        Gradient Clipping Threshold     & -                                     & 0.5                                    \\ 
        Bandwidth                       & \multicolumn{2}{c}{100 KHz}                                   \\ 
        Transmission Power                       & \multicolumn{2}{c}{1 mW}                                   \\ 
        \bottomrule
    \end{tabularx}
\end{table}

We conducted our experiments using the Sentiment140 dataset \cite{go2009twitter}, which contains 1.6 million tweets. To adapt the dataset for resource-constrained devices, only the text and target labels (0 for negative sentiment and 1 for positive sentiment) were retained, and the dataset size was halved. All experimental configurations are summarized in Table~\ref{table:experiment_params}.
Binary Phase Shift Keying (BPSK) is the digital modulation scheme adopted to transmit the quantized data. Energy consumption and CO\textsubscript{2} emissions were monitored using the Eco2AI framework \cite{budennyy2022eco2ai}, with measurements taken every 10 seconds during the FL cycles.
In the CL setup, 3 users collaborated in sending the data to the server, and the SL setup involved one user and one server. Both CL and FL models were trained for 50 cycles using the same batch size across all experiments. In SL, the model was partitioned at a specific split layer $L$, where convolutional and pooling layers were executed on the user's device to minimize computational complexity, while the rest of the layers were handled by the server. The update rules for the SGD optimizer are defined as
\begin{equation}
v_{t+1} = \mu \cdot v_t + \eta \cdot \nabla \mathcal{L}(w_t),
\end{equation}
\begin{equation}
w_{t+1} = w_t - v_{t+1}.
\end{equation}
To ensure stable training and prevent exploding gradients, gradient clipping was applied with a threshold of $\tau = 0.5$. If the gradient norm exceeded $\tau$, it was scaled down to maintain the norm at or below this threshold as shown in Algorithm.~\ref{alg:efficient_split_learning}. Gradient clipping stabilized the update steps, preventing large parameter changes that could destabilize convergence.

{To evaluate the privacy we used an autoencoder under the same experimental setup to measure reconstruction error. We introduced an adversary for each user and then averaged the reconstruction errors across users. For simplicity, the autoencoder was trained on the same dataset with direct access to the raw inputs and followed the same setup in Table~\ref{table:experiment_params}; in practical scenarios, where attackers lack such access, the reconstruction task would be substantially more challenging.}

\begin{figure*}[h]
    \centering
    \captionsetup[subfigure]{justification=centering}
    
    \begin{subfigure}[b]{0.42\textwidth}
        \centering
        \includegraphics[width=\linewidth, trim={0 0 0 1cm}, clip]{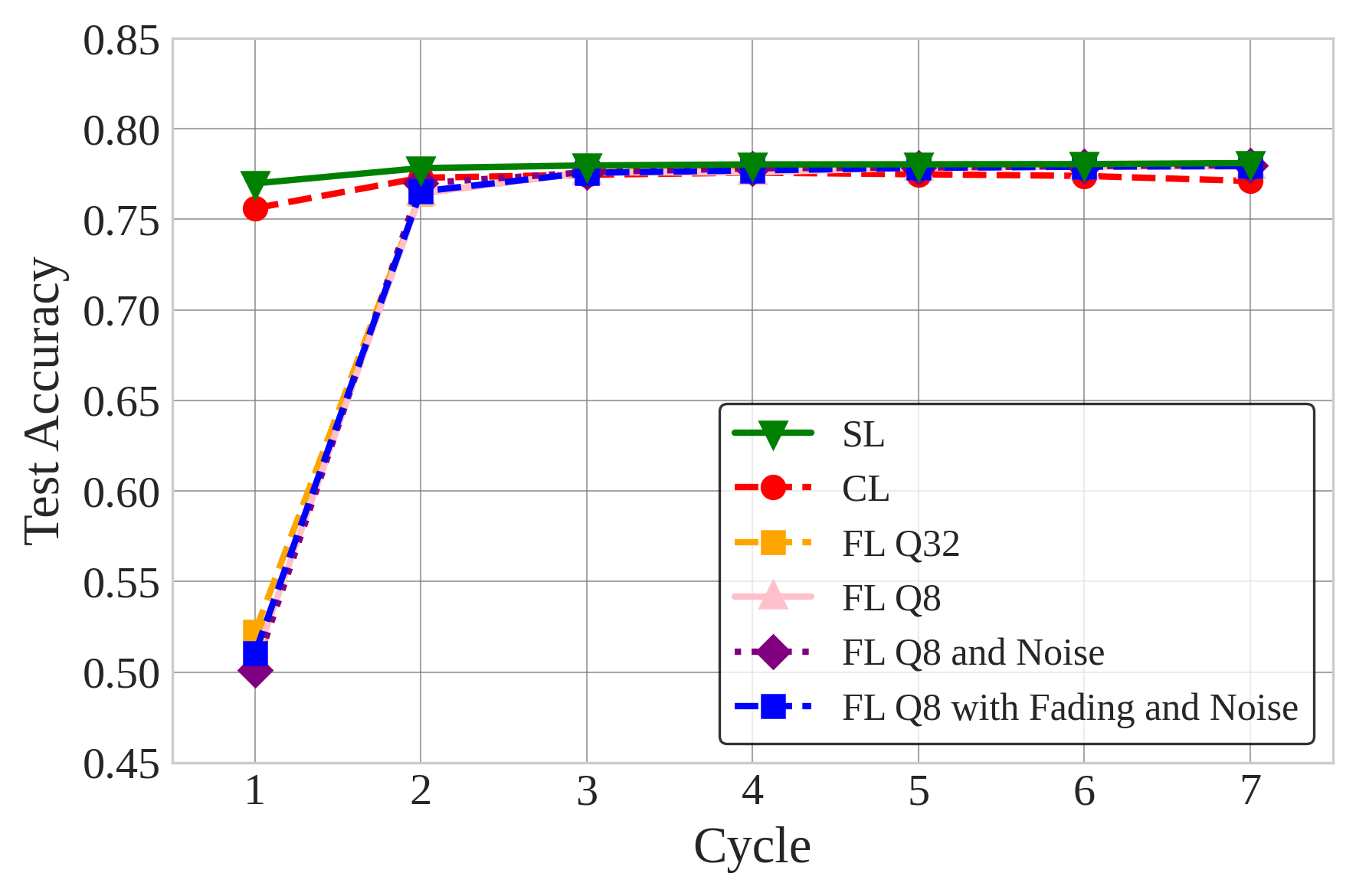}
        \caption{Centralized, Federated, and Split Learning\\Comparison}
        \label{fig:general_methods_results}
    \end{subfigure}
    \begin{subfigure}[b]{0.42\textwidth}
        \centering
        \includegraphics[width=\linewidth, trim={0 0 0 1cm}, clip]{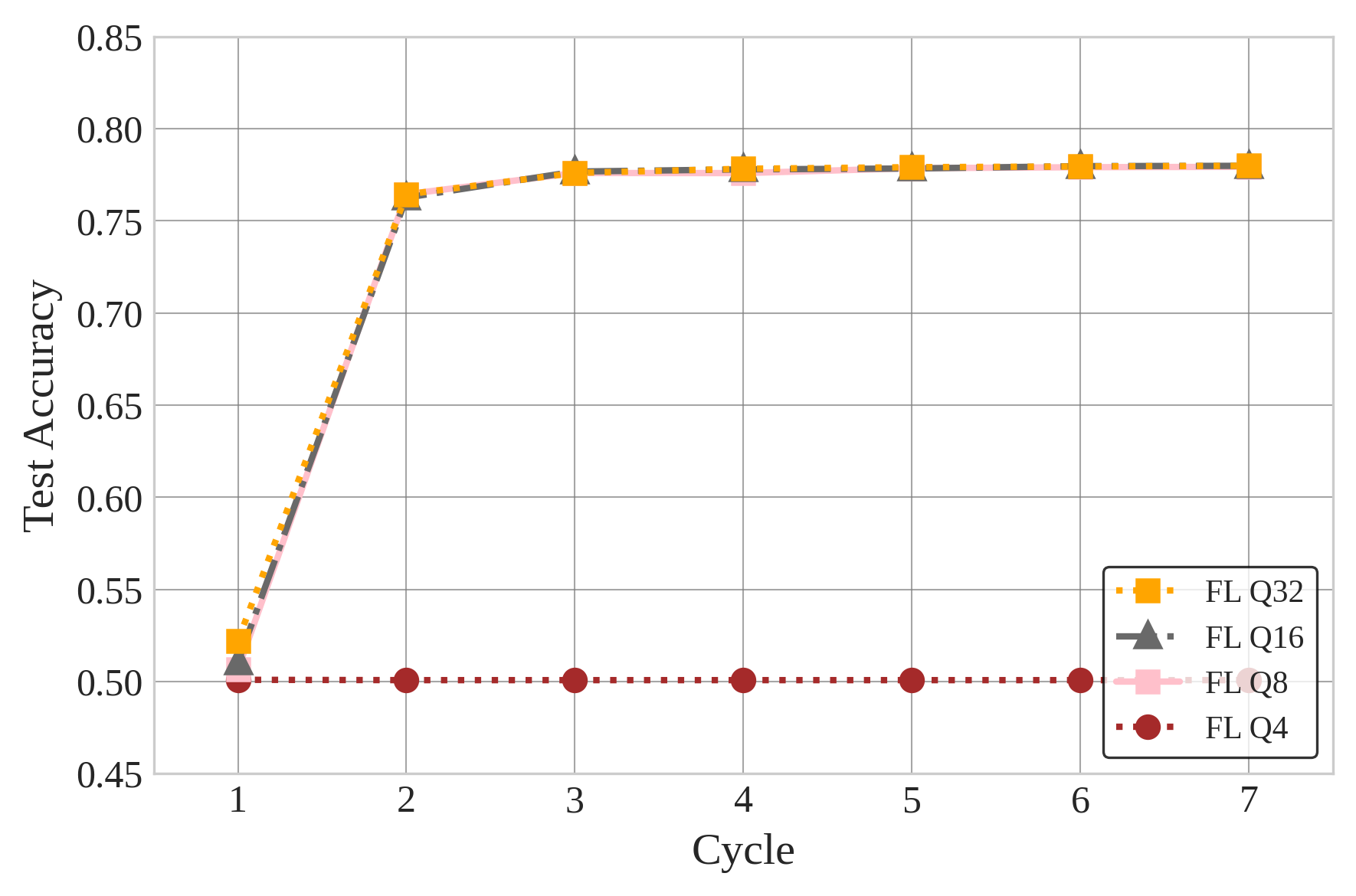}
        \caption{Accuracy vs. Cycle for Different\\Quantization Levels (Q4, Q8, Q16, Q32)}
        \label{fig:quantization_results}
    \end{subfigure}
    
    \vspace{0pt}
    
    \begin{subfigure}[b]{0.42\textwidth}
        \centering
        \includegraphics[width=\linewidth, trim={0 0 0 1cm}, clip]{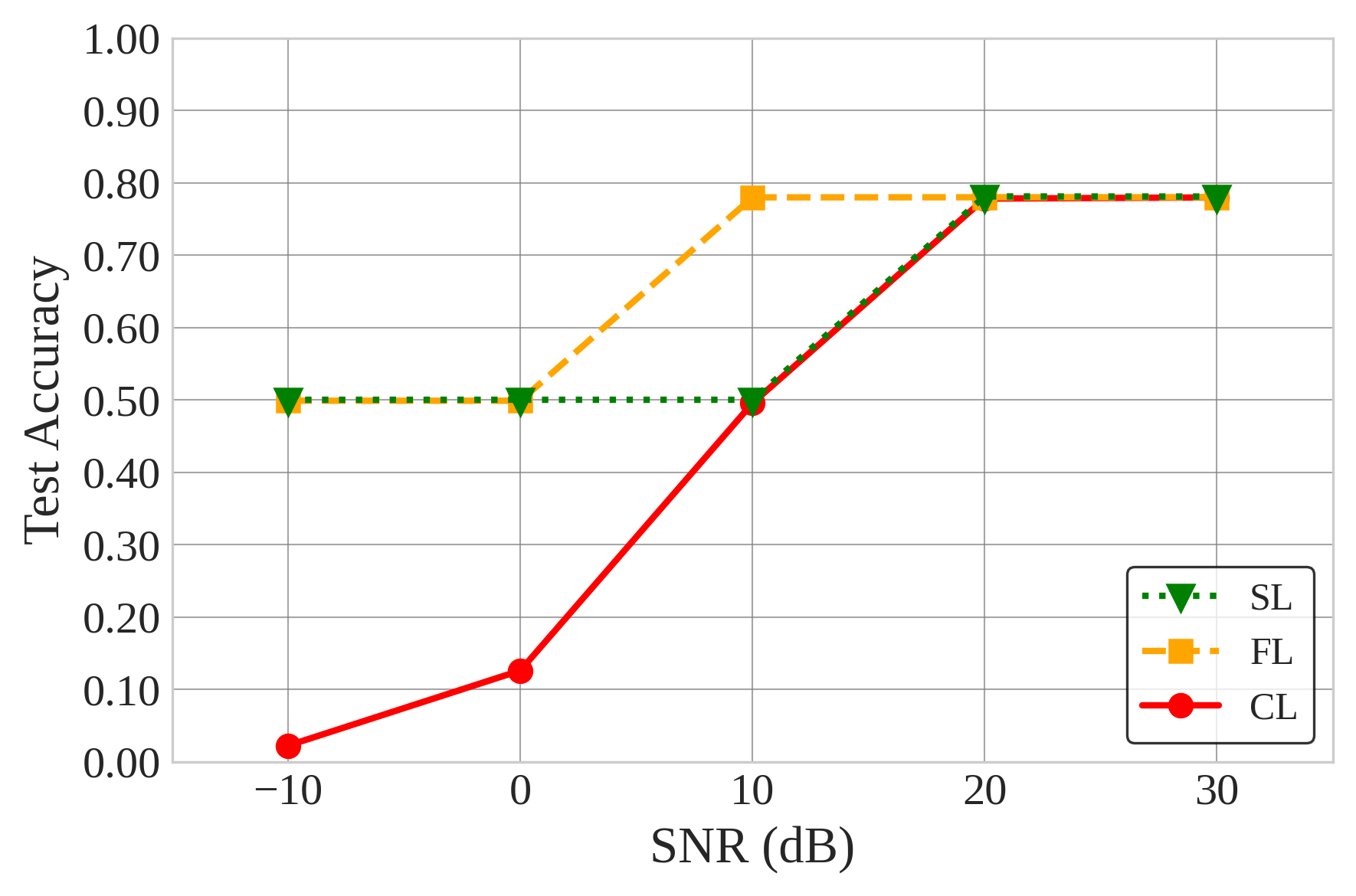}
        \caption{Accuracy vs. SNR for different Learning methods}
        \label{fig:SNR_results}
    \end{subfigure}
    \begin{subfigure}[b]{0.42 \textwidth}
        \centering
        \includegraphics[width=\linewidth, trim={0 0 0 1cm}, clip]{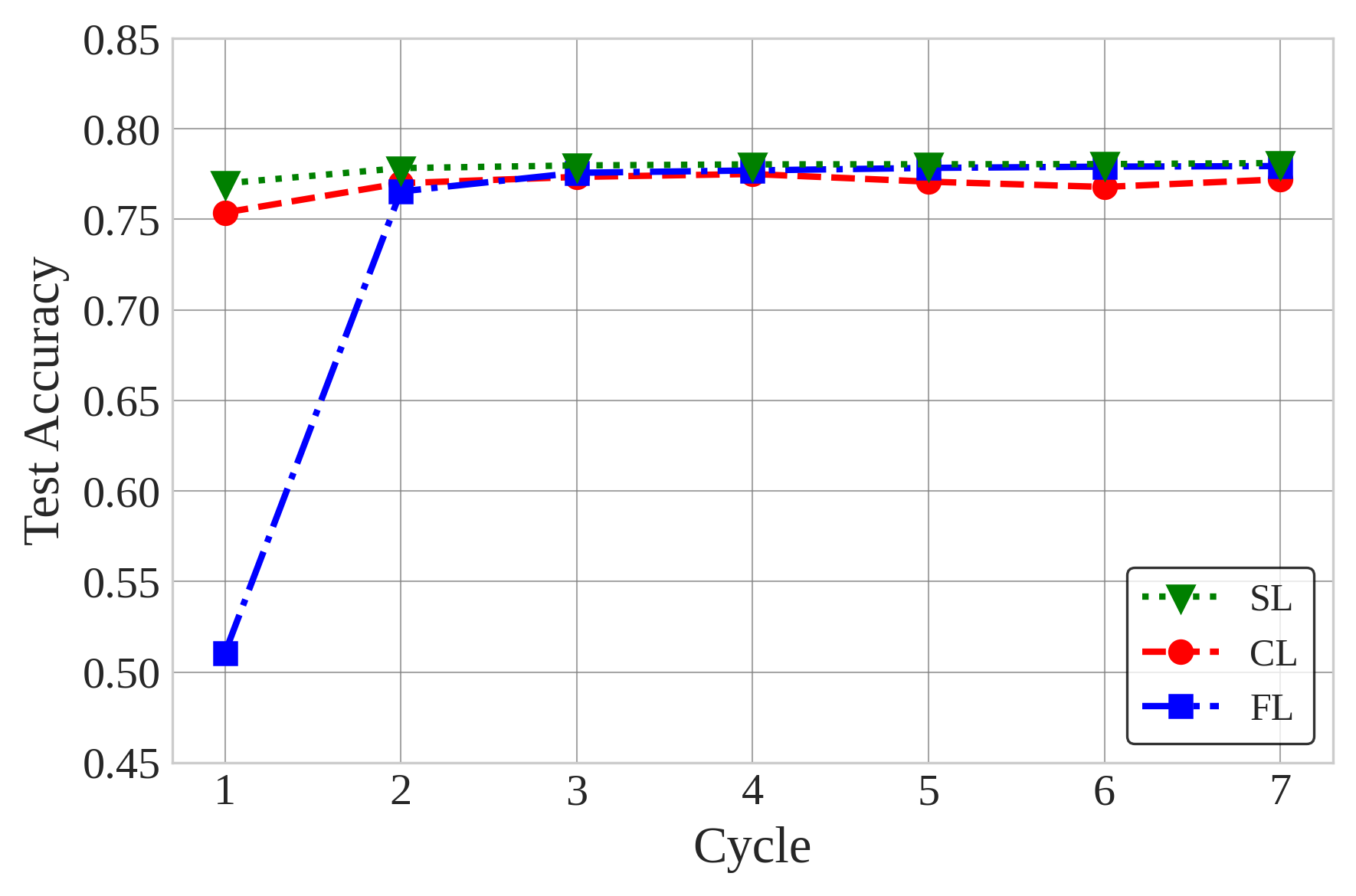}
        \caption{Accuracy vs. Cycle with Fading and Noise}
        \label{fig:fading_noise_results}
    \end{subfigure}
    
    \caption{\small{Comparative analysis of learning methods: (a) centralized, federated, and split learning comparison; 
    (b) Accuracy vs. cycle for different quantizations; (c) Accuracy vs. SNR in different learning methods; 
    (d) Accuracy vs. cycle fading and noise.}}
    \label{fig:combined_results}
\end{figure*}

\subsection{Model Architecture}
\label{modelArch} 

The model, with a total of 89,673 parameters and an approximate size of 175.14 KB using 16-bit precision, is tailored for resource-constrained environments. This compact design, combined with our dual-approach framework, offers significant benefits in resource utilization and privacy protection. {A compression encoder factoring by four is adopted to economize the use of resources.}

\subsubsection{Federated learning}
The FL architecture consists of several layers: an input layer that accepts fixed-length sequences of maximum length \( N \), an embedding layer that transforms input sequences into dense vectors of size 8, a convolutional layer with 32 filters and a kernel size of 3. 
After that comes  an LSTM layer with 32 units to capture temporal dependencies in the sequential output, a dense layer with 16 units using ReLU activation and L2 regularization, and finally, an output layer with 1 unit using sigmoid activation for binary classification.

\subsubsection{Split learning}
In our SL model, the user-side computation is represented by the function \( f_u(x) \), where \( x \) is the input data processed through the initial layers of the model, including convolutional and pooling layers. The output of \( f_u(x) \) is then sent to {encoder for compression then sent to }the server, where the server-side computation \( f_s(f_u(x)) = y \) {takes input and passes it to a decoder for decompression then }processes the data further using the LSTM and subsequent layers to produce the final output \( y \). This split allows the server to handle the more computationally intensive tasks.

\subsection{Experimental Results}

We evaluate and compare learning frameworks based on model accuracy, computational energy, and communication energy while also examining the effects of noise and Rayleigh fading.
Fig.\ref{fig:general_methods_results} shows that both CL and FL, with 8-bit (Q8) and 32-bit (Q32) quantization, converge at approximately 0.78 accuracy. However, introducing noise with an SNR level of 20 dB and Rayleigh fading in CL, as shown in Fig.\ref{fig:fading_noise_results}, leads to a slight degradation in accuracy.
This degradation occurs because CL transmits raw data, which is directly affected by noise and fading during transmission. In contrast, FL transmits quantized model weights, which are smaller and more structured, making them less susceptible to transmission impairments.
As shown in Fig.~\ref{fig:quantization_results}, lower quantization levels (e.g., Q4) reduce accuracy due to precision loss, while Q8 and higher offer a favorable trade-off between accuracy and communication efficiency. Q8 emerges as the optimal balance.

\begin{table*}[h!]
\centering
\caption{Performance comparison of decentralized learning algorithms: average metrics over 10 runs}
\label{table:energy_costs}
\small
\begin{tabular}{@{}lccccccc@{}}
\toprule
\textbf{Algorithm} & \textbf{Total Bits} & \textbf{Accuracy} & \textbf{Recon. Error} & \textbf{Comp. Energy} & \textbf{Comm. Energy} & \textbf{Total Energy} \\
& (M bits) & & & (J) & (J) & (J) \\
\midrule
Central          & 115.7 & 0.7803 & 0.0154 & 0 & {0.3459} & {0.3459} \\
FL Q8           & \textbf{0.72} & \textbf{0.7806} & 0.0671 & 60.8200 & \textbf{0.0021} & 60.8221 \\
SL (Early Cut)  & {2580.48} & {0.7800} & \textbf{0.2681} & \textbf{3.4512} & 7.7162 & \textbf{11.1674} \\
\bottomrule
\end{tabular}
\begin{flushleft}
\footnotesize
\textit{Note:} Total Bits are reported per user. For all entries, computational energy is reported on the user side. \vspace{-2mm}
\end{flushleft}
\end{table*}

In Fig.~\ref{fig:SNR_results}, we analyze the effect of varying SNR on model accuracy across FL, CL, and the proposed SL framework. As SNR increases, accuracy improves, especially between 0 dB and 10 dB, with FL demonstrating the highest overall performance. Beyond 20 dB, accuracy plateaus at about 0.78 across all methods, with 20 dB identified as a reasonable balance between accuracy and communication power. Again, FL is more robust in noisy and fading environments. Meanwhile, Fig.~\ref{fig:fading_noise_results} shows the effect of Rayleigh fading and noise together. We set the SNR to 20 dB and observe that both FL with Q8 and SL maintain high accuracy, demonstrating robustness under challenging real-world conditions, despite communication channel imperfections.

Table \ref{table:energy_costs} compares the three approaches. The evaluation reveals key insights about privacy-computation tradeoffs. The proposed SL scheme demonstrates the highest reconstruction error of 0.2681, indicating stronger privacy preservation as it becomes increasingly difficult to recover raw data from intermediate activations $S_i$. In contrast, FL and CL show concerning privacy vulnerabilities with low reconstruction errors of 0.0671 and 0.0154, respectively.

For user-side computational efficiency, SL and FL were evaluated at 20 dB SNR, while CL shows zero user-side energy in the table. However, CL’s substantial server-side energy—though not shown—should be considered.
SL requires the least computational energy, reaching 3.45 J, compared to FL 60.82 J, but incurs the highest communication energy due to frequent intermediate activation transfers, even with compression. FL balances between communication and computation, making it suitable for communication-limited scenarios.
In both FL and SL, user-side computation dominates energy use, with SL achieving the lowest overall user energy in TinyML settings. SL also offers superior efficiency, emitting 10 times less CO\textsubscript{2} than FL and 20 times less than CL, as measured by Eco2AI. These results support SL as an ideal choice for TinyML NLP classification.

\section{Conclusion}
We explored TinyML approaches for semantic text sentiment classification via FL and SL. Our approaches are designed to be energy-efficient and privacy-preserving alternatives to CL. It is clear that quantization techniques such as 8-bit quantization appeared to be optimum in our scenario.
SL excels in reducing user-side computation and CO\textsubscript{2}  emissions while maintaining accuracy in noisy conditions but incurs higher communication energy. FL, especially with Q8 quantization, offers an optimal balance between computational efficiency, communication cost, and data privacy, making it ideal for decentralized environments with limited bandwidth. The proposed semantic TinyML SL and FL approaches are considered highly efficient for tiny devices with limited energy resources, where SL is the most privacy-preserving, energy-efficient and environment-friendly. Future work could extend these TinyML schemes to LLMs and integrate differential privacy to further enhance communication efficiency and security.

\section*{Acknowledgments} \vspace{0mm}
This work is supported by the KAUST Office of Sponsored Research under Award ORA-CRG2021-4695.

\bibliographystyle{IEEEtran}  
\bibliography{references}   
\vspace{12pt}
\color{red}
\end{document}